# Extracting periodontitis diagnosis in clinical notes with RoBERTa and regular expression

Yao-Shun Chuang
School of Biomedical Informatics
University of Texas Health
Science Center at Houston,
Houston, Texas, USA
yao-shun.chuang@uth.tmc.edu

Chun-Teh Lee
Deaprtment of Periodontics and
Dental Hygiene
The University of Texas Health
Science Center at Houston
School of Dentistry
Houston, Texas, USA
Chun-Teh.Lee@uth.tmc.edu

Ryan Brandon
Department of Oral Health
Sciences
Temple University Kornberg
School of Dentistry
Philadelphia, Pennsylvania, USA
ryan.brandon@uth.tmc.edu

Trung Duong Tran
Diagnostic and Biomedical
Sciences
The University of Texas Health
Science Center at Houston
School of Dentistry
Houston, Texas, USA
trung.duong.tran@uth.tmc.edu

Oluwabunmi Tokede
Oral Healthcare Quality and
Safety
The University of Texas Health
Science Center at Houston
School of Dentistry
Houston, Texas, USA
oluwabunmi.tokede@uth.tmc.edu

Muhammad F. Walji
Diagnostic and Biomedical
Sciences
University of Texas Health
Science Center at Houston,
Houston, Texas, USA
Muhammad.F.Walji@uth.tmc.edu

Xiaoqian Jiang
School of Biomedical Informatics
University of Texas Health
Science Center at Houston,
Houston, Texas, USA
xiaoqian.jiang@uth.tmc.edu

*Abstract*— This study aimed to utilize text processing and natural language processing (NLP) models to mine clinical notes for the diagnosis of periodontitis and to evaluate the performance of a named entity recognition (NER) model on different regular expression (RE) methods. Two complexity levels of RE methods were used to extract and generate the training data. The SpaCy package and RoBERTa transformer models were used to build the NER model and evaluate its performance with the manual-labeled gold standards. The comparison of the RE methods with the gold standard showed that as the complexity increased in the RE algorithms, the F1 score increased from 0.3-0.4 to around 0.9. The NER models demonstrated excellent predictions, with the simple RE method showing 0.84-0.92 in the evaluation metrics, and the advanced and combined RE method demonstrating 0.95-0.99 in the evaluation. This study provided an example of the benefit of combining NER methods and NLP models in extracting target information from free-text to structured data and fulfilling the need for missing diagnoses from unstructured notes.

*Keywords— Natural language processing, Named entity recognition, Regular expression, Transformer, Missing diagnosis, Dentistry*

## I. Introduction

Electronic Health Records (EHRs) are patients' clinical notes in digital form and contain personal contact information, medical history, laboratory results, and treatment plans. Although the original purpose of EHRs was for billing reasons, EHRs allow for the organization and analysis of incalculable amounts of patients' structured and unstructured data [1]. This unintended benefit of EHRs could improve the efficiency and effectiveness of the healthcare system. Further benefits of EHRs include process quality control, guideline compliance, and patient outcome monitoring [1]. However, a significant challenge to the benefits of big data is that much of the information is recorded in free-text unstructured format. This makes it difficult to mine for analysis, as the data is not organized in a way that can be easily understood or processed. In order to make use of the data, it must first be organized and structured in a way that allows for analysis.

While procedures and lab results are commonly structured, clinicians regularly document medical history, examination findings and operative notes in narrative text [2]. This is especially true in dentistry, where despite the evolution of diagnostic standards and coding terminologies [3], not many cases are documented with an accurate diagnosis, recorded in a structured format. Dental clinicians often find themselves in situations where they may complete a dental procedure without selecting or writing a proper diagnosis. This can be due to a number of reasons, such as a lack of attention or no need for insurance claim. This is a major concern for quality patient care, as it can lead to improper documentation of patient health records. Without a proper diagnosis, future care providers may not be able to accurately assess the patient's condition, and may not be able to provide the best possible treatment. Comprehensive and accurate medical records are critical for improving the continuity of care and patient safety [4]. Furthermore, the lack of a proper diagnosis for a patient can lead to delays in insurance coverage for a procedure or even a denial of coverage. However, manually searching, summarizing, or statistically analyzing the huge amount of data related to this issue is difficult and time-consuming. Natural language processing and text data mining repressent state-of-the-art solutions to this problem.

Text data mining is a powerful tool for extracting information and discovering knowledge from huge amounts of noisy, incomplete, or vague data [5]. In human language data, this process could be subdivided into the fields of Natural language processing (NLP) and text-mining, which use computer algorithms and programs to interpret and analyze natural language or regular expression [6]. These techniques are important for discovering and summarizing valuable information. For example, as of January 2023 on PubMed's website, it stated that it contained over 35 million articles [7]. With the aid of NLP and text-mining approaches, these abundant journals and papers may be able to provide fruitful results and insights [8]. Many researchers have already demonstrated that these methods could produce remarkable results in many fields, including text classification, sentiment analysis, and summarization [9]. Thus, named entity recognition (NER) is one of the techniques in NLP to solve the difficulties of information extraction from EHRs by mining structured data from free text.

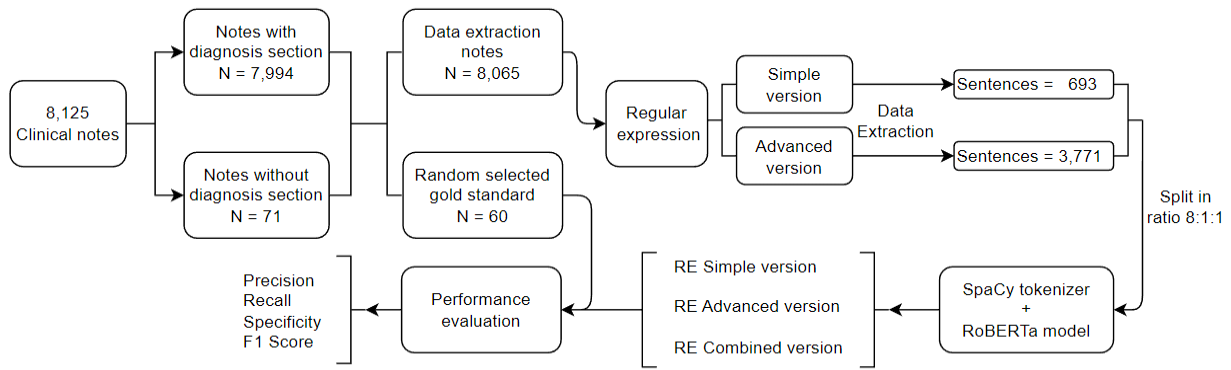

Fig. 1. A flowchart of this study from data collection, filtering, initial labeling with regular expression and advanced NLP model for named entity recognition.

Periodontitis affecting almost half of the adults aged 30 or older in the United States is an inflammatory disease characterized by gingival inflammation and alveolar bone loss around teeth [10]. In 2018, a new classification of periodontal diseases was introduced and redefined diagnoses of periodontitis by a multidimensional staging and grading system [3]. Staging, extent, and grading are three key terms used to diagnose periodontitis. Staging is determined by the severity of the disease at the time of presentation, as well as the complexity of the disease management. Extent represents the percentage of periodontitis-affected teeth at the identified stage. Grading is determined by the risk of disease progression associated with history of disease progression, local and systemic factors. As these diagnostic terms are relatively new to dental care providers, it is common to find that providers do not write a proper and structured diagnosis of periodontitis in their notes. To address this issue, NER methods can be used to identify diagnoses of periodontitis in clinical notes, even when there are unstructured diagnostic terms or missing diagnoses associated with clinical procedures. The objective of this study is to combine text processing and advanced NLP models to mine clinical notes for periodontal diagnoses in order to help fulfill the missing diagnosis problem in an efficient manner. By utilizing text processing and NLP models, this paper seeks to provide an effective way to extract and analyze the relevant information from clinical notes. To do this, we investigated the performance of the NER model on two different approaches of regular expression methods. This study explored the tradeoffs between model performance and initial human efforts to obtain labels. This research provided an efficient and accurate way to mine clinical notes for periodontal diagnoses, allowing researchers to gain a better understanding of the effectiveness of the NER model.

## II. METHODS AND MATERIAL

### A. Dataset

In this study, the data were extracted from EHR for the period January 1 2021 through December 31, 2021. Cases were included based on having an examination visit with complete periodontal charting including pocket depths, clinical attachment loss and free gingival margin to the cemento-enamel junction. The cases were limited to the age over 16 with a minimum of 10 natural teeth present, and having recent bitewing radiographs (within 6 months of the examination). Based on the criteria, there were a total of 5,495 qualifying patients in the dataset. For each of these cases, the clinical notes documented within one month of the examination were extracted, totaling 8,125 clinical notes in all. Further, to check the accuracy from the model predictions, 60 clinical notes were randomly selected as the gold standard and manually labeled by an examiner (an experienced dentist) who did not write any of the notes. These notes were excluded from the training dataset in both RE approaches. Thus, a total of 8,065 clinical notes were used for generating the training data.

### B. Target Information

Following the American Academy of Periodontology (AAP)/ European Federation of Periodontology (EFP) 2018 classification of periodontitis [11], clinical notes of patients diagnosed with periodontitis were utilized. There are four levels of Stage (I, II, III, and IV), three levels of Grade (A, B, C) and three types of Extent (localized, generalized, molar/incisor pattern). The extent, molar /incisor pattern, was not considered in this study due to its rareness.

### C. Data Extraction

Fig. 1 was a flowchart of this study. The regular expression (RE) was utilized to extract the key information for generating the data for model training. The clinical notes were first split by a next-line character to separate each section after the contents were reviewed. Given clinical notes of patients enrolled in UTHealth Houston dental clinics have a specific format, the data would be screened in the section "D" or "Diagnosis" with ignoring the letter of cases and found 98.7% of patients' records contained this target section. Then, RE further proceeded to every section with two approaches, simple and advanced versions. The simple version checked the labels in the order of diagnosis section, extent, stage, grade, and the word "periodontitis" after screening over 20 notes, where each corresponding label had to be existed and in the correct order. The advanced RE version neglected the label order presented in the records and allow partial labels which means not require all label must be existed and provide more flexibility. Additionally, both versions were addressed with the typo for the Extent label and the word "periodontitis". Table I showed 5 examples and the results from both RE approaches, where the "O" indicated that it was able to be captured with that RE method, and vice versa for "X". While the fifth example should have been gathered using the advanced version RE, a filter was employed to verify the presence of the Stage and Grade, along with the Diagnosis, in the sentences to ensure that the sentences contained the desired labels. The further explanation was in the section of pre-processing and post-processing. The partial RE of simple and advanced versions were in Equation (1) and (2), respectively. After applied with both RE approaches, only 693 sentences were found in the simple RE and 3,771 sentences in the advanced RE. To fine-tune the model, the results from both versions were split in the ratio of 8:1:1. The detail numbers were in Table II.

"^(?<DX>(d|diagnosis)) ?(?<DX_I>[:-]+) ?(?<DDF_EXTENT>gener?a?l?i?z?e?d?|local?i?z?e?d?) ?(?<STAGE_STR>stage) (?<STAGE_VAL>[\w\d]*)"   (1)

"^(?=.*(d|diagnosis) ?([:-]+) ?)(?=.*(gener?a?l?i?z?e?d?|local?i?z?e?d?))?(?=.*(stage) ([\w\d]*))?"   (2)

TABLE I.    EXAMPLES FOR THE RESULTS FROM RE APPROACHES.

| No | Example sentence | Simple RE | Advanced RE |
|---|---|---|---|
| 1 | d: generalized stage iii grade c periodontitis. | O | O |
| 2 | d- localized periodontitis, stage 3 grade b. | X | O |
| 3 | d: tentative diagnosis is stage 3 grade c generalized | X | O |
| 4 | d- stage iii grade b periodontitis. | X | O |
| 5 | d : generalized plaque induced gingivitis | X | X |

TABLE II.    DISTRIBUTION OF DATASET AND TRAINING DATA.

| RE Type | Total collected records | Training set | Validation set | Testing set |
|---|---|---|---|---|
| Simple RE | 693 | 554 | 69 | 70 |
| Advanced RE | 3,771 | 3,016 | 377 | 378 |

### D. Pre-processing and post-processing

Data pre-processing and post-processing were further applied in this study. In the data pre-processing, after the labels were extracted from RE approaches, this dataset was further filtered by checking the existence of the three labels, Stage, Grade, and Diagnosis. Such a procedure could make sure the target information was in the diagnosis section and contained Stage and Grade information. The post-processing was implemented after model generated the results. The data might be extracted from the notes correctly but not usable for evaluation. So, data generalization was applied, including correcting the typo of Extent and generalizing the Stage and Grade values. In addition, some values of the label Grade would be captured with symbols, such as dot, coma, and other symbols. The example was No.2 in Table I, where the Grade would be captured as "b.". Thus, these symbols would be removed in the post-processing. Also, the value would be empty if the data could not be generalized. Additionally, because multiple diagnoses may exist in one clinical note, the most severity diagnosis was selected by comparing the order of severity in the hierarchy as Stage, Extent, and then Grade.

### E. Methods of NLP modeling and framework

The spaCy package was utilized as a tokenizer and proceeded for NER tasks because the spaCy tokenization uses a no-destructive approach and keeps all whitespace and punctuation, which allows the extracting data to reconstruct and further save into spaCy training data format [12]. In addition, its architecture allows users to not only customize the NLP pipeline but also provide high performance. Further, RoBERTa base can be easily applied in NER models for this research.

RoBERTa stands for Robustly Optimized BERT Pre-training approach and is the same architecture as Bidirectional Encoder Representations from Transformers (BERT). The difference between them is that the pretrained process of RoBERTa is applied a dynamic masked language modeling to avoid over-memorizing the training dataset [13]. Since the target information extraction in this study doesn't contain many medical professional terms, RoBERTa-base model was selected for this general-domain task due to performing better in these general-purpose tasks than BERT model [14]. In the following paper, the simple RE model would represent the model trained by the simple version of RE and the similar meaning for the advanced RE model, while the simple RE method would consider as the approach using the simple version of regular expression and vice versa for the term of the advanced RE method. The combined version of the model utilized the advanced RE version as the foundation for its results, which were subsequently enhanced with the simple version by supplementing any missing information from the advanced version. Only two notes required substitution using the advanced RE.

### F. Evaluation metrics

A confusion matrix was used to evaluate NER performance. True positive (TP) values represent the disease status of periodontitis in the gold standards and are correctly recognized by the NER model, and vice versa in the true negative (TN). False positive (FP) values indicate the results from NER model incorrectly predict the status of diagnosis. False negative (FN) reveals the NER model misses the actual diagnosis of periodontitis. In addition, six metrics were utilized to evaluate the algorithm's performance. Precision (P) is known as positive predictive value; recall also known as true positive rate or sensitivity; specificity is testing the ratio of TN and; F1 score is the harmonic mean of precision and recall, where the equation is respectively in Equation (3), (4), (5), and (6). Furthermore, due to the uncertainty and imbalance of disease status distribution in periodontitis, the macro average and weighted average were applied in the three metrics above. The macro average is calculated by simply averaging the evaluation values in that fields, and the weighted average is generated by weighting the evaluation values by corresponding quantity, which is in Equation (7). $W$ represents as weighted average and $w_i$ is the number of quantity for that particular label; n is the number of total labels, and $Xi$ is the evaluation value.

$$Precision = \frac{TP}{(TP+FP)} \quad (3)$$

$$Recall/Sensitivity = \frac{TP}{(TP+FN)} \quad (4)$$

$$Specificity = \frac{TN}{(TN+FP)} \quad (5)$$

$$F1\ score = \frac{2 \times Precision \times Recall}{Precision+Recall} \quad (6)$$

$$W = \frac{\sum_{i=0}^{n} w_i X_i}{\sum_{i=0}^{n} w_i} \quad (7)$$

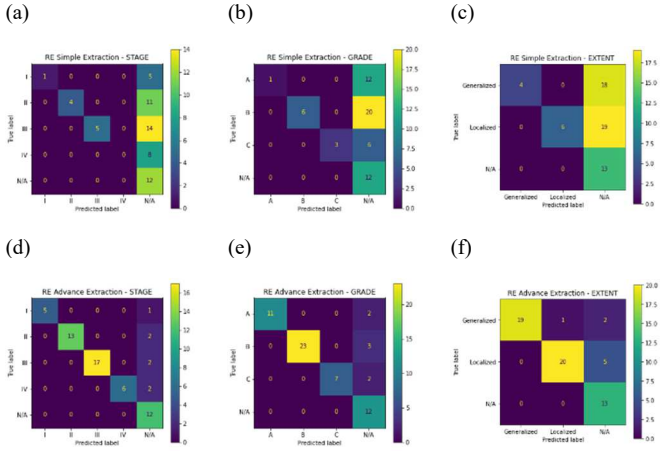

Fig. 2. Confusion matrix of the gold standards along regular expression

## III. RESULTS

In this study, the dataset was applied to two RE methods and further utilized the extracted data for training with RoBERTa model, then compared with the 60 gold standard clinical notes, which an examiner manually labeled. The confusion matrix results from two RE methods are in Fig. 2. Table III revealed the results of evaluation metrics of two RE approaches compared with the gold standard. The precision in Stage, Grade, and Extent from the simple RE approach was around 0.7, 0.8, and 0.8, respectively, while the overall precisions were above 0.9 in the advanced RE. Both recall and F1 scores were near 0.4 in the simple RE and the advanced RE was near 0.9. For the model prediction, Fig. 3 revealed the confusion matrix results of the NER model predictions from two RE and the combined approaches. The evaluation metrics of the model prediction with the gold standards are shown in Table IV. The results of the simple RE model showed a precision of 0.94, and recalls and F1 scores of around 0.87 in both macro and weighted averages. The advanced RE model had an overall performance of near 0.98 in Stage and Grade labels, except 0.95 in the Extent label. The combined approach saw a Stage label of 1.0 in the evaluation metrics, and a slight improvement of around 0.02 in the Extent label, while the Grade label maintained the same performance. The specificity results for all three labels were around 0.94-1.0, except the simple RE approach had results of around 0.74-0.84. These results are shown in Fig. 4.

TABLE III. EVALUATION METRICS COMPARISON BETWEEN THE GOLD STANDARD AND REGULAR EXPRESSION APPROACHES

|  |  | Precision | | Recall | | F1 score | |
| --- | --- | --- | --- | --- | --- | --- | --- |
|  |  | Simple | Advanced | Simple | Advanced | Simple | Advanced |
| Stage | Macro average | 0.65 | 0.93 | 0.34 | 0.87 | 0.30 | 0.88 |
|  | Weighted average | 0.71 | 0.93 | 0.37 | 0.88 | 0.34 | 0.89 |
| Grade | Macro average | 0.81 | 0.91 | 0.41 | 0.88 | 0.35 | 0.88 |
|  | Weighted average | 0.85 | 0.93 | 0.38 | 0.88 | 0.35 | 0.89 |
| Extent | Macro average | 0.75 | 0.87 | 0.47 | 0.89 | 0.37 | 0.86 |
|  | Weighted average | 0.84 | 0.90 | 0.38 | 0.87 | 0.36 | 0.87 |

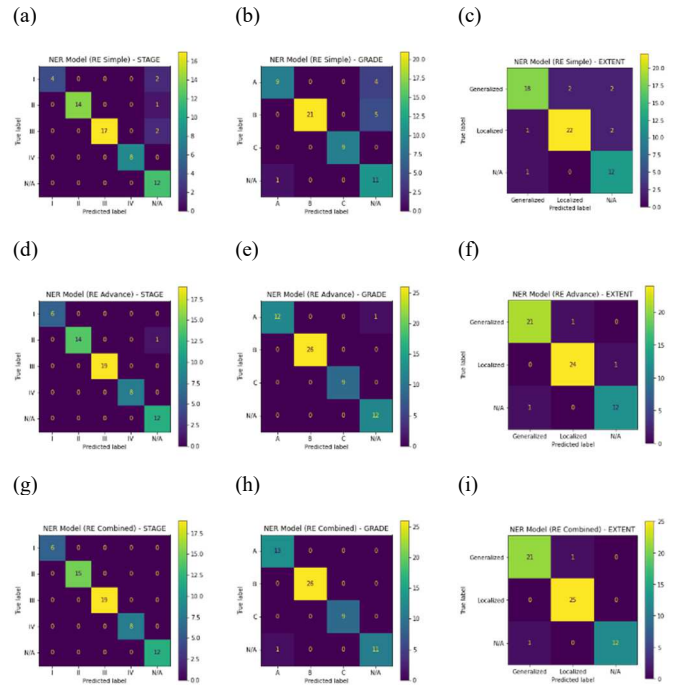

Fig. 3. Confusion matrix of the gold standards with NER model predictions

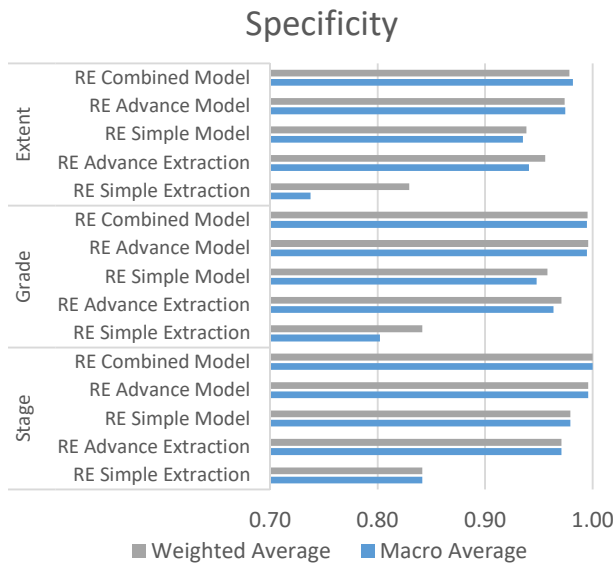

Fig. 4. The macro and weighted average of specificity over all approaches.

TABLE IV. THE COMPARISON BETWEEN NER MODEL TRAINED BY THE SIMPLE AND ADVANCED VERSIONS OF REGULAR EXPRESSION WITH THE GOLD STANDARDS AS WELL AS THE COMBINED APPROACH.

|  |  | Precision | | | Recall | | | F1 score | | |
|---|---|---|---|---|---|---|---|---|---|---|
|  |  | Simple | Advanced | Combined | Simple | Advanced | Combined | Simple | Advanced | Combined |
| Stage | Macro average | 0.94 | 0.98 | 1.00 | 0.90 | 0.99 | 1.00 | 0.91 | 0.99 | 1.00 |
|  | Weighted average | 0.94 | 0.98 | 1.00 | 0.92 | 0.98 | 1.00 | 0.92 | 0.98 | 1.00 |
| Grade | Macro average | 0.86 | 0.98 | 0.98 | 0.85 | 0.98 | 0.98 | 0.84 | 0.98 | 0.98 |
|  | Weighted average | 0.89 | 0.98 | 0.98 | 0.83 | 0.98 | 0.98 | 0.84 | 0.98 | 0.98 |
| Extent | Macro average | 0.86 | 0.95 | 0.97 | 0.87 | 0.95 | 0.96 | 0.86 | 0.95 | 0.96 |
|  | Weighted average | 0.87 | 0.95 | 0.97 | 0.87 | 0.95 | 0.97 | 0.87 | 0.95 | 0.97 |

algorithms can lead to more accurate and comprehensive results due to the model having more comprehensive training and superior prediction results. However, this complexity results in a positive correlation with the time spent. NLP models provide a great solution to this issue. During error analysis, more than 10 cases in the 60 gold standard notes showed that although the models could successfully capture the target information, it still included some symbols such as dots, commas, and other symbols. This caused errors in the evaluation metrics and thus, post-processing using rule-based text processing is necessary after using the model's prediction. Additionally, the majority of recall metrics are lower than precision, which is a desired result as the goal is to extract the missed diagnosis. Post-processing is also necessary for data standardization, as the NER model correctly captured some target terms with typos and informal terms, such as "Grade B" being written as "Grade ii" and other incorrect values, thus requiring post-processing to fix these issues. The limitations of the NER model were found through error analysis, one of the limitations being that the NER model could never be perfect. Through the evaluation tables, both simple and advanced RE models provided great results, but they were not perfect, leading to the generation of a combined RE model.

## IV. DISCUSSION

This study found that a NER model could extract periodontitis diagnosis with high accuracy after fine-tuning with a seed. It demonstrated the potential of NER models to solve real-world problems, even with simple algorithms and small amounts of training data. Although a comprehensive regular expression could produce a similar outcome to a simple RE model, the NER model was able to produce outstanding predictions. It was able to take unstructured notes and turn them into a structured format, fulfilling the need for missing diagnoses. This study showed that not only periodontitis but other dental diseases could be implemented in the model to create more complete and comprehensive structured data in EHRs for further clinical use.

The strengths of the learning ability of NER models were demonstrated through both RE approaches, which are the current general approaches used. A common issue of high false negatives in the confusion matrix in RE results indicated that the contents of clinical notes contained high diversity, making it difficult to capture all the rules. Creating more complex RE

By combining the results from the simple and advanced RE models, the simple model was more flexible in certain cases and was able to cover the shortcomings of the advanced model. This led to the Stage reaching an F1 score of 1.0 and the Extent reaching scores of 0.96-0.97. However, the Grade label remained the same, as there was one note where the NER model was unable to capture the information. Additionally, the complexity of free-text notes posed a limitation, as the diagnosis was written in another two notes in a different format with other explanations, which included the target terms but not for periodontitis diagnosis. Other minor limitations included the fact that several components could be replaced with other packages or models to test the performance, such as the NER model, which could be replaced with RoBERTa, BERT, ClinicalBERT, or other large language models.

## V. CONCLUSION

The need to capture missing diagnoses associated with clinical procedures from notes in the dental field can be fulfilled with the use of Natural Language Processing models such as the NER model. Clinical notes are generated by humans and can contain a variety of terms and formats, making it impossible to use regular expression (RE) alone to meet the goal. We found a good combination strategy. By using RE methods to create the training data, the RoBERTa model can learn from the patterns and provide accurate predictions. This model also makes up for the limitation of only using RE approaches. Moving forward, our model will undergo testing with dental datasets from other institutions. Furthermore, there is potential for this model to be utilized in the diagnosis of other illnesses and expanded to various medical fields. Additionally, by increasing the complexity and including rare medical terminologies, it may be feasible to integrate other large language models, such as GPT, into the production pipeline for optimal results.

ACKNOWLEDGMENT

XJ is CPRIT Scholar in Cancer Research (RR180012), and he was supported in part by Christopher Sarofim Family Professorship, UT Stars award, UTHealth startup, the National Institute of Health (NIH) under award number R01AG066749, R01LM013712, and U01TR002062, and the National Science Foundation (NSF) #2124789

REFERENCES

[1] S. Yanamadala, D. Morrison, C. Curtin, K. McDonald, and T. Hernandez-Boussard, "Electronic Health Records and Quality of Care," *Medicine (Baltimore)*, vol. 95, no. 19, p. e3332, May 2016, doi: 10.1097/MD.0000000000003332.

[2] D. Demner-Fushman, W. W. Chapman, and C. J. McDonald, "What can natural language processing do for clinical decision support?," *J. Biomed. Inform.*, vol. 42, no. 5, pp. 760–772, Oct. 2009, doi: 10.1016/j.jbi.2009.08.007.

[3] P. N. Papapanou et al., "Periodontitis: Consensus report of workgroup 2 of the 2017 World Workshop on the Classification of Periodontal and Peri-Implant Diseases and Conditions," *J. Periodontol.*, vol. 89, no. S1, pp. S173–S182, 2018, doi: 10.1002/JPER.17-0721.

[4] J. S. Patel, "Utilizing Electronic Dental Record Data to Track Periodontal Disease Change," Jul. 2020, doi: 10.7912/C2/958.

[5] W.-T. Wu et al., "Data mining in clinical big data: the frequently used databases, steps, and methodological models," *Mil. Med. Res.*, vol. 8, no. 1, p. 44, Aug. 2021, doi: 10.1186/s40779-021-00338-z.

[6] Z. Zeng, H. Shi, Y. Wu, and Z. Hong, "Survey of natural language processing techniques in bioinformatics," *Comput. Math. Methods Med.*, vol. 2015, 2015.

[7] "About," *PubMed*. https://pubmed.ncbi.nlm.nih.gov/about/ (accessed Jan. 28, 2023).

[8] J. Lee et al., "BioBERT: a pre-trained biomedical language representation model for biomedical text mining," *Bioinformatics*, vol. 36, no. 4, pp. 1234–1240, Feb. 2020, doi: 10.1093/bioinformatics/btz682.

[9] A. Borjali, M. Magnéli, D. Shin, H. Malchau, O. K. Muratoglu, and K. M. Varadarajan, "Natural language processing with deep learning for medical adverse event detection from free-text medical narratives: A case study of detecting total hip replacement dislocation," *Comput. Biol. Med.*, vol. 129, p. 104140, Feb. 2021, doi: 10.1016/j.compbiomed.2020.104140.

[10] P. I. Eke, G. O. Thornton-Evans, L. Wei, W. S. Borgnakke, B. A. Dye, and R. J. Genco, "Periodontitis in US Adults: National Health and Nutrition Examination Survey 2009-2014," *J. Am. Dent. Assoc. 1939*, vol. 149, no. 7, pp. 576-588.e6, Jul. 2018, doi: 10.1016/j.adaj.2018.04.023.

[11] K. S. Kornman and P. N. Papapanou, "Clinical application of the new classification of periodontal diseases: Ground rules, clarifications and 'gray zones,'" *J. Periodontol.*, vol. 91, no. 3, pp. 352–360, 2020, doi: 10.1002/JPER.19-0557.

[12] H. Eyre et al., "Launching into clinical space with medspaCy: a new clinical text processing toolkit in Python," *AMIA. Annu. Symp. Proc.*, vol. 2021, pp. 438–447, Feb. 2022.

[13] P. Lewis, M. Ott, J. Du, and V. Stoyanov, "Pretrained Language Models for Biomedical and Clinical Tasks: Understanding and Extending the State-of-the-Art," in *Proceedings of the 3rd Clinical Natural Language Processing Workshop*, Online, Nov. 2020, pp. 146–157. doi: 10.18653/v1/2020.clinicalnlp-1.17.

[14] Y. Liu et al., "RoBERTa: A Robustly Optimized BERT Pretraining Approach." arXiv, Jul. 26, 2019. doi: 10.48550/arXiv.1907.11692.